\documentclass[letterpaper]{article} %
\usepackage[preprint]{aaai2027}  %
\usepackage[hyphens]{url}  %
\usepackage{graphicx} %
\usepackage{natbib}  %
\usepackage{caption} %
\usepackage{amsmath}
\usepackage{amssymb}
\usepackage{amsfonts}
\usepackage{amsthm}
\usepackage{mathtools}
\usepackage{bm}

\usepackage{algorithm}
\usepackage{algpseudocode}

\usepackage{booktabs}
\usepackage{tabularx}
\usepackage{multirow}
\usepackage{makecell}
\usepackage{array}
\usepackage{subcaption}
\usepackage{tikz}
\usetikzlibrary{arrows.meta,positioning,fit,calc,shapes.geometric}

\usepackage{siunitx}
\usepackage{xcolor}
\usepackage{enumitem}

\usepackage{newfloat}
\usepackage{listings}
\DeclareCaptionStyle{ruled}{labelfont=normalfont,labelsep=colon,strut=off}
\floatstyle{ruled}
\newfloat{listing}{tb}{lst}{}
\floatname{listing}{Listing}

\usepackage{cleveref}

\makeatletter
\renewcommand{\theHALG@line}{\thealgorithm.\arabic{ALG@line}}
\makeatother

\setlist{nosep,leftmargin=*}
\providecommand{\href}[2]{#2}

\definecolor{DataBlue}{HTML}{EAF2F8}
\definecolor{LearnPurple}{HTML}{F1ECF7}
\definecolor{PhaseBlue}{HTML}{E8F1FA}
\definecolor{PhaseGreen}{HTML}{EDF5E9}
\definecolor{PhaseGold}{HTML}{FCF4DF}
\definecolor{TransformGray}{HTML}{F5F5F3}
\definecolor{TraceGold}{HTML}{FBF7E8}
\definecolor{RobotGreen}{HTML}{EEF5EA}
\definecolor{RuleBlue}{HTML}{EDF3F8}
\definecolor{LineGray}{HTML}{6E7378}

\title{Phase-Conditioned Action Contracts for Auditable Microsurgical Control}

\author{
  Jiaxi You\textsuperscript{1}, Yiwei Shi\textsuperscript{1},
  Yanlin Li\textsuperscript{2}, Rihui Song\textsuperscript{2},\\
  Weibing Li\textsuperscript{2}, Kai Huang\textsuperscript{2},
  Weiru Liu\textsuperscript{1}\thanks{Corresponding author.}
}
\affiliations{
  \textsuperscript{1}University of Bristol, United Kingdom\\
  \textsuperscript{2}Sun Yat-sen University, China
}

\begin{document}

\maketitle
\begin{abstract}
Staged robotic procedures may reuse a fixed-dimensional policy
output even though the admissible motions and physical meanings of
its coordinates change across phases. We call this
\emph{action-semantic shift}. To address it, we propose a
phase-conditioned executable policy that separates learned action
proposal from deterministic command formation. Phase-specific heads
generate proposals, while an action contract removes inadmissible
components, enforces state-dependent bounds, and decodes the
remaining action into phase-dependent task-space commands. We further
introduce failure-focused local repair and a read-only provenance
trace for exact command replay and rule-level auditing. Across three held-out retinal-injection evaluation blocks, our method
achieves \(94.27\pm0.64\%\) procedural completion and
\(64.40\pm2.43\%\) success within \(35\,\mu\mathrm{m}\), with no hard violations. Applying the contract and decoder to the same frozen
Correction network raises completion from \(12.27\%\) to
\(90.47\%\). The trace exactly reconstructs 96,350 commands, and all
1,996 binding interventions follow their declared effects. These
results show the importance of explicit executable action
semantics for auditable staged microsurgical control.
\end{abstract}

\section{Introduction}

Many procedural control tasks require a policy to act across stages whose admissible motions differ. Microsurgical retinal injection is a particularly sharp example. During pre-aim, the needle must translate and rotate to approach a prescribed entry waypoint and align with a planned insertion axis. Once the current needle line is accepted, however, the line becomes a geometric commitment: guarded entry and insertion should remain on the locked axis, eliminating lateral and angular freedom. Near the subretinal target, even the remaining axial feed must be restricted to avoid overshoot. The central challenge is therefore not only to predict an accurate action value, but also to preserve the physical meaning and admissible range of each action channel as the procedural phase changes. This matters in microsurgical control because small phase-inappropriate motions can turn an otherwise accurate learned action into an unsafe or invalid physical command.

A common way to learn such controllers is to use task-space imitation learning. A policy observes the current task geometry, procedural state, and safety-related variables, and maps them to a fixed-dimensional action. This approach is attractive because expert demonstrations can provide direct supervision for closed-loop behavior, while task-space actions avoid learning low-level joint commands tied to a specific robot embodiment \citep{Argall2009,Osa2018}. Phase flags and safety variables can be included in the observation so that the network can infer which stage it is in. Learner-state correction methods, such as DAgger-style aggregation, further improve robustness by querying expert labels on states reached by the learner rather than only on states from the original demonstrations \citep{Ross2011}. These methods are widely used because they provide a simple and scalable recipe: keep one policy interface, enrich the observation, and expand the supervised dataset when distribution shift appears.

However, this mainstream formulation leaves a fundamental limitation unresolved. Treating phase as an input condition does not by itself define how the output channels should be interpreted in each phase. In retinal injection, the same five-dimensional policy output cannot have the same physical meaning throughout the procedure: before entry, the channels represent translational and angular correction; after line acceptance, only axial feed should remain active; near the target, even this feed must be capped or stopped. Learner-state correction can broaden state coverage, but it still trains a policy whose action contract is implicit if the output mapping is shared across stages. The root cause is an action-representation mismatch rather than only a state-distribution mismatch. We refer to this phenomenon as \emph{action-semantic shift}: as a staged procedure advances, procedural phase changes not only which action is desirable, but also which action components are physically meaningful and admissible.

We address this problem by making procedural phase part of an executable
command-generation contract. A shared encoder produces a latent state,
hard-routed Phase-A, A2, and B heads generate bounded neural proposals, and
the selected proposal is transformed by terminal feed regulation, a
phase-conditioned action contract, and a deterministic physical decoder. In
Phase A, all five channels remain available for Cartesian and angular
correction. In A2 and B, the contract sets non-feed channels exactly to zero,
enforcing locked-axis motion after line acceptance. The A2 contract
lower-bounds forward progress during guarded entry, while the Phase-B contract
and decoder progressively restrict feed and produce zero forward motion inside
the terminal stop band. A read-only command-provenance trace records the
selected head, raw and regulated proposals, phase contract, and decoded
task-space command, enabling exact reconstruction and local rule-level audits
of the actual command path. Across three paired held-out retinal-injection
blocks of 500 geometries each, the complete structured policy reaches
$94.27\pm0.64\%$ Flow and $64.40\pm2.43\%$ success within
35~$\mu$m, compared with $1.20\pm0.35\%$ Flow and
$1.20\pm0.35\%$ success within 35~$\mu$m for monolithic behavioral cloning,
and $12.27\pm1.45\%$ Flow and $11.73\pm1.15\%$ success within
35~$\mu$m for monolithic Correction IL. In a separately retrained matched
structural family, the full configuration reaches $99.73\pm0.12\%$ Flow and
$75.00\pm3.14\%$ success within 35~$\mu$m; removing phase-conditioned action
structure reduces these rates to $24.40\pm0.72\%$ and $0.00\pm0.00\%$,
respectively, and produces a $0.07\pm0.12\%$ hard-violation rate, whereas the
full structured family produces none. The trace audit reconstructs all 96,350
official policy actions exactly, and all 1,996 eligible one-rule-off
interventions follow their predeclared local direction.

The work makes three contributions:
1) We formulate staged microsurgical control as an action-semantic shift problem, where procedural phase changes the physical meaning and admissible set of policy output channels.2) We introduce an executable phase-conditioned action contract that combines hard-routed neural proposals, terminal feed regulation, locked-axis admissibility, and deterministic task-space decoding. 3) We evaluate the resulting policy through full-task comparisons, matched structural ablations, command-level trace reconstruction, controlled one-rule interventions, and independent one-robot-at-a-time FK/IK compatibility analysis of frozen task-space outcomes.

\section{Related Work}
\textbf{Robot-assisted microsurgery and retinal needle control.}
Actively stabilized handheld tools, motorized force-sensing microneedles, and robot-assisted retinal cannulation have demonstrated the value of mechanical precision in delicate ocular procedures \citep{MacLachlan2012,Gonenc2017,Gijbels2018}. More recent work studies autonomous needle navigation in ex vivo porcine eyes \citep{Zhang2023}, deep-learning-based retinal vein cannulation \citep{Zhang2025}, and autonomous targeted retinal injection across phantoms and animal models \citep{Bian2026}. This literature emphasizes sensing, planning, force interaction, and robotic execution. We isolate a complementary question: how should a learned task-space command change its semantics as a staged microsurgical procedure advances?

\textbf{Task-space control and action representations.}
Operational-space control specifies motion or force objectives in task coordinates and realizes them through manipulator kinematics and dynamics \citep{Khatib1987}. In robot learning, action-space design can trade learning efficiency against control of redundant configurations \citep{Mazzaglia2024}. Our policy remains in needle task space, where the clinically meaningful variables are entry geometry, needle tip, axis, and insertion depth. Robot joints are introduced only in a separate compatibility protocol, preventing embodiment-specific IK error from being conflated with policy inference.

\textbf{Imitation learning and learner-state correction.}
Behavioral cloning can fail under covariate shift because small errors move the learner into states poorly represented by demonstrations \citep{Argall2009,Osa2018}. DAgger-style aggregation queries expert labels on learner-reached states \citep{Ross2011}. We use phase-labeled analytic supervision to establish a broad anchor, then perform a narrower second stage: failed A-to-A2 handoffs are reconstructed under the frozen anchor, only the Phase-A head is updated, and successful Phase-A and A2/B outputs are retained toward the anchor.

\textbf{Safety filters and faithful command-level evidence.}
Control barrier functions can enforce conditions associated with forward invariance \citep{Ames2019}, while shielding can replace actions that violate a temporal-logic specification \citep{Alshiekh2018}. Our method does not append a separate online shield. Instead, phase-dependent admissibility is built into the action contract and decoder. Interpretability research emphasizes matching evidence to its intended use \citep{DoshiVelez2018} and warns that post-hoc explanations may be unfaithful in high-stakes settings \citep{Rudin2019}. Local surrogates such as LIME approximate predictions in a neighborhood \citep{Ribeiro2016}; by contrast, our trace records the actual deterministic refinement of each issued command and is tested by exact replay and controlled local interventions.

\section{Method}
\label{sec:method}

\subsection{Phase-Dependent Action Semantics}
\label{sec:action-semantics}

We consider a staged control process with observation
$o_t\in\mathcal{O}$, procedural phase $\phi_t\in\Phi$, and
physical task-space command $s_t\in\mathcal{S}$. The phase
$\phi_t$ is supplied by the task logic and identifies the current
procedural stage; it is not predicted by the policy. The learned controller represents neural proposals and
contract-compliant normalized actions in a shared,
fixed-dimensional normalized coordinate space $\mathcal{B}\subseteq\mathbb{R}^{m}$,
where $m$ is the coordinate dimension. Importantly,
$\mathcal{B}$ is an ambient coordinate space rather than an
admissible action set. A neural proposal and an executable
normalized action can therefore have the same dimension while
having different operational roles and different admissibility
status.

Specifically, raw and regulated proposals are vectors in
$\mathcal{B}$ but are not necessarily admissible for execution.
For each phase $\phi$ and observation $o$, we define the
state-dependent admissible normalized action set
$\mathcal{A}_{\phi}(o)\subseteq\mathcal{B}$.

An executable normalized action must belong to
$\mathcal{A}_{\phi}(o)$. A fixed coordinate dimension therefore does not imply fixed
action semantics. Different phases may permit different motion
components, impose different directional or magnitude restrictions,
or assign the same normalized coordinate a different physical
interpretation. We refer to variation in
$\mathcal{A}_{\phi}(o)$ or in its physical interpretation across
phases as \emph{action-semantic shift}. Providing phase indicators as observation features can help a
network represent the current stage, but does not by itself determine
which predicted components are permitted to reach the physical
system. We therefore make phase part of the executable
command-generation path rather than treating it only as contextual
input.

\subsection{Phase-Conditioned Executable Policy}
\label{sec:executable-policy}

The controller separates learned proposal generation from
deterministic command formation. A frozen preprocessing map
$N:\mathcal{O}\rightarrow\widetilde{\mathcal{O}}$ produces
$\widetilde{o}_t=N(o_t)$, and a shared encoder computes
$z_t=f_{\theta_e}(\widetilde{o}_t)$. For each phase $k\in\Phi$, a phase-indexed head returns a bounded
neural proposal $ b_t^k=h_{\theta_k}(z_t)\in\mathcal{B}$. Here, $k$ is a dummy index ranging over the available procedural
phases, whereas $\phi_t$ denotes the current phase supplied by the
task logic. Hard routing selects exactly one proposal:
\begin{equation}
    b_t
    =     \sum_{k\in\Phi}
    \mathbb{I}[\phi_t=k]\,b_t^k
    =     b_t^{\phi_t}
    \in\mathcal{B}.
    \label{eq:phase-routing-main}
\end{equation}

All heads may be evaluated for logging, but only
$b_t^{\phi_t}$ enters command formation. The model is therefore
one policy with a shared representation and hard phase routing,
rather than a runtime ensemble of independently executed
controllers.

Observation conditioning and executable routing have distinct
roles. Phase features may inform the shared representation, whereas
Equation~\eqref{eq:phase-routing-main} explicitly determines which
phase-specific proposal mapping is permitted to enter the executed
command path.

A state-dependent regulator acts on the selected proposal before
the deterministic contract. For each phase, it has the form
$R_{\phi}:     \mathcal{O}\times\mathcal{B}
    \rightarrow     \mathcal{B},     \bar b_t     =     R_{\phi_t}(o_t,b_t)$.
Thus, $b_t$ is the raw phase-routed proposal and $\bar b_t$ is
the regulated proposal. Both are vectors in $\mathcal{B}$. The regulator is the identity outside designated operating regions.
When required by the task, it may also be directionally restricted,
for example so that a risk-sensitive progress coordinate can be
reduced but not enlarged. Any learned parameters of
$R_{\phi}$ are included in the overall learned parameter set
$\theta$.

Although regulation may improve the proposal using the current
state, $\bar b_t$ is not yet guaranteed to belong to the admissible
action set $\mathcal{A}_{\phi_t}(o_t)$. Admissibility is instead
guaranteed by the deterministic action contract introduced next.

\textbf{Executable action contract.} For each phase, the task specifies a binary channel mask
$\mu_{\phi}(o)\in\{0,1\}^{m}$ and componentwise lower and upper
bounds $
    \ell_{\phi}(o),u_{\phi}(o)\in\mathbb{R}^{m},
    \ell_{\phi}(o)\preceq u_{\phi}(o)$,
where $\preceq$ denotes componentwise inequality. For every
disabled coordinate $j$, the declared admissible value is zero:
$\mu_{\phi,j}(o)=0 \quad\Longrightarrow\quad \ell_{\phi,j}(o)=u_{\phi,j}(o)=0$. The resulting admissible normalized action set is $ \mathcal{A}_{\phi}(o)
    = \Bigl\{
        a\in\mathcal{B}
        \ \Big|\
        \bigl(\mathbf{1}-\mu_{\phi}(o)\bigr)\odot a=0,
        \ell_{\phi}(o)\preceq a\preceq u_{\phi}(o)
    \Bigr\}$, where $a$ is an $m$-dimensional normalized action vector and $\odot$ denotes elementwise multiplication. In other words,
$\mathcal{A}_{\phi}(o)$ is a set of admissible vectors, rather than
a single action. For a fixed observation $o$ and phase $\phi$, the executable
contract is a  mapping
$C_{\phi}(o,\cdot):
    \mathcal{B}
    \rightarrow
    \mathcal{A}_{\phi}(o)$.

It maps the regulated proposal into the final contract-compliant
normalized action:
    $a_t
    =
    C_{\phi_t}(o_t,\bar b_t)
    =
    \mu_{\phi_t}(o_t)\odot
    \operatorname{clip}\!\left(
        \bar b_t,
        \ell_{\phi_t}(o_t),
        u_{\phi_t}(o_t)
    \right)
    \in
    \mathcal{A}_{\phi_t}(o_t)$,
where clipping is applied componentwise.

The mask removes phase-inappropriate degrees of freedom, while the
bounds encode directional, progress, and terminal restrictions.
These masks and bounds are supplied by the task instantiation rather
than inferred implicitly from demonstrations.

Consequently, for every raw neural proposal $b\in\mathcal{B}$,
the complete regulator--contract composition produces an admissible
normalized action:
$
    C_{\phi}\!\left(
        o,
        R_{\phi}(o,b)
    \right)
    \in
    \mathcal{A}_{\phi}(o),
    \forall b\in\mathcal{B}$.
Thus, declared phase-wise admissibility does not depend on the
neural policy learning the constraints from data.

\textbf{Phase-dependent physical decoding.} The contract operates in normalized coordinates. A
phase-dependent decoder assigns the contract-compliant normalized
action its physical task-space meaning:
$s_t = D_{\phi_t}(o_t,a_t)$,
where
$D_{\phi}:
    \left\{
        (o,a)\in
        \mathcal{O}\times\mathcal{B}
        \mid
        a\in\mathcal{A}_{\phi}(o)
    \right\}
    \rightarrow
    \mathcal{S}$.
The decoder may use phase-specific reference frames, task geometry,
geometric commitments created at earlier transitions, or state
variables such as remaining progress. Different phases can
therefore share the same normalized coordinate dimension while
assigning those coordinates different physical meanings.

The complete command-generation chain is
$
\underbrace{b_t\in\mathcal{B}}_{\text{raw neural proposal}}
    \longrightarrow
    \underbrace{\bar b_t\in\mathcal{B}}_{\text{regulated proposal}}
    \longrightarrow
    \underbrace{
        a_t\in\mathcal{A}_{\phi_t}(o_t)
    }_{\text{admissible normalized action}}
    \longrightarrow
\underbrace{s_t\in\mathcal{S}}_{\text{physical command}}$. The distinction between these quantities is therefore operational,
rather than dimensional. The vectors $b_t$, $\bar b_t$, and $a_t$
share the normalized coordinate space $\mathcal{B}$, but only
$a_t$ is guaranteed to satisfy the phase-dependent admissibility
conditions. The decoded quantity $s_t$ belongs to the physical
task-space command space $\mathcal{S}$. Hard routing isolates proposal mappings associated with different
phases; the regulator provides designated state-dependent proposal
adjustments; the contract removes inadmissible components and
enforces declared bounds; and the decoder assigns the resulting
normalized action its current physical meaning. A phase transition
therefore changes not only the information available to the policy,
but also the executable action set and its physical interpretation. \textbf{Retinal-injection instantiation.} For the retinal-injection benchmark, the phase set is
$ \Phi = \{
        \mathrm{A},
        \mathrm{A2},
        \mathrm{B} \}$.

During pre-aim (Phase $\mathrm{A}$), translational and
orientational correction channels remain available. Once the
current needle line is accepted, guarded entry
(Phase $\mathrm{A2}$) and insertion (Phase $\mathrm{B}$) remove
lateral and angular channels and interpret the remaining active
coordinate as progress along the locked line. Phase $\mathrm{A2}$ introduces a lower progress bound, whereas
Phase $\mathrm{B}$ uses a remaining-depth-dependent upper bound
and a terminal no-forward-motion region. The
implementation-specific architecture, contract constants, and
decoder equations are fixed in the frozen benchmark configuration
used for all reported experiments.

\subsection{Failure-Focused Phase-Local Repair}
\label{sec:phase-local-repair}

The executable contract resolves ambiguity in action admissibility
and physical interpretation, but it does not guarantee an accurate
proposal at every learner-reached state. We address this remaining
state-coverage error through failure-focused phase-local repair.

\textbf{Anchor learning.} The first stage uses an offline, phase-labeled dataset
$\mathcal{D}_0
    =
    \left\{
        (o_i,\phi_i,y_i^E)
    \right\}_{i=1}^{N_0}$,
where
$y_i^E\in\mathcal{B}$
is an expert target for the regulated, pre-contract proposal. The
target is expressed in the same normalized, phase-dependent
coordinate semantics as the regulated proposal.

In our notation, $y_i^E$ is not the contract output $a_i$ and is
not the decoded physical command $s_i$. It supervises the learned
proposal-generation path before the deterministic contract and
physical decoder are applied. To avoid introducing a separate, ambiguous policy symbol, we write
the regulated proposal produced for an arbitrary observation--phase
pair as
$\bar b_{\theta}(o,\phi)
    =
    R_{\phi}\!\left(
        o,
        h_{\theta_{\phi}}\!\left(
            f_{\theta_e}(N(o))
        \right)
    \right)
    \in
    \mathcal{B}$.
Here, $o$ is the observation, $\phi$ is the procedural phase, and
$\theta$ denotes the collection of all learned parameters,
including encoder, phase-head, and regulator parameters when the
regulator is learned.

All learned modules are first optimized using
$\mathcal{L}_{\mathrm{anchor}}
    =
    \frac{1}{N_0}
    \sum_{i=1}^{N_0}
    \ell\!\left(
        \bar b_{\theta}(o_i,\phi_i),
        y_i^E
    \right)$,
where $\ell(\cdot,\cdot)$ is instantiated as mean-squared error in
our experiments. The resulting anchor parameters are denoted by
$\theta^0$.

The deterministic contract $C_{\phi}$ and physical decoder
$D_{\phi}$ remain outside this regression objective. They are
applied only after the regulated proposal has been produced.

\textbf{Failure-state reconstruction.} The frozen anchor is then executed in closed loop without online
expert intervention. Let
$\Phi_r\subseteq\Phi$
denote the set of procedural phases selected for repair. States associated with a predeclared repairable failure class are
selected from the anchor traces and reconstructed from the same task
seed and initial condition. A reconstructed state is retained only
when its replayed regulated proposal matches the stored regulated
proposal within a fixed numerical tolerance. The expert then
provides a corrected pre-contract target for each accepted state,
forming $\mathcal{D}_{\mathrm{corr}}$. Successful anchor rollouts provide a retention set
$\mathcal{D}_{\mathrm{ret}}$. For each retained state $(o,\phi)$,
the target is the frozen anchor proposal
$ y^0 = \bar b_{\theta^0}(o,\phi)
    \in \mathcal{B}$.

Correction examples are therefore supervised by the expert,
whereas retention examples preserve successful anchor behavior.
No observation is paired with a target generated from a different
trajectory state.

\textbf{Phase-local update.} Let $\theta_r$ denote the parameters of the proposal heads
associated with the phases in $\Phi_r$. During repair, only
$\theta_r$ is trainable; the shared encoder, heads outside
$\Phi_r$, regulator, and all other learned modules remain frozen at
their anchor values.

The correction and retention losses are
\begin{align}
    \mathcal{L}_{\mathrm{corr}}
    &=
    \mathbb{E}_{
        (o,\phi,y^E)
        \sim
        \mathcal{D}_{\mathrm{corr}}
    }
    \left[
        \ell\!\left(
            \bar b_{\theta}(o,\phi),
            y^E
        \right)
    \right],
    \label{eq:correction-loss-main}
    \\
    \mathcal{L}_{\mathrm{ret}}
    &=
    \mathbb{E}_{
        (o,\phi,y^0)
        \sim
        \mathcal{D}_{\mathrm{ret}}
    }
    \left[
        \ell\!\left(
            \bar b_{\theta}(o,\phi),
            y^0
        \right)
    \right].
    \label{eq:retention-loss-main}
\end{align}

The repair objective is
$ \theta_r^{\star}
    =
    \arg\min_{\theta_r}
    \left[
        \mathcal{L}_{\mathrm{corr}}
        +
        \lambda_{\mathrm{ret}}
        \mathcal{L}_{\mathrm{ret}}
    \right],
    \lambda_{\mathrm{ret}}\geq 0$.
The correction term improves selected learner-reached failures,
while the retention term limits deviation on successful states from
the repaired phases. Because every learned module on a non-repaired
routed path remains frozen, its regulated proposal is unchanged for
the same fixed observation:
\begin{equation}
    \bar b_{\theta^{\star}}(o,\phi)
    =
    \bar b_{\theta^0}(o,\phi),
    \phi\notin\Phi_r.
    \label{eq:non-repair-preservation-main}
\end{equation}

Equation~\eqref{eq:non-repair-preservation-main} is a
fixed-observation mapping statement. It does not imply
downstream trajectories, because an earlier repaired action may
change which states are subsequently visited. For the retinal-injection benchmark,
$\Phi_r = \{\mathrm{A}\}$,
where $\mathrm{A}$ denotes the pre-aim procedural phase. The
upright phase label $\mathrm{A}$ is distinct from the calligraphic
admissible-action-set notation $\mathcal{A}_{\phi}(o)$. Only the pre-aim proposal head is updated; the shared encoder,
guarded-entry head, insertion head, and terminal regulator remain
fixed. Detailed state-selection rules, replay tolerances, dataset
sizes, optimization settings, and candidate promotion checks are
fixed before evaluation and recorded with the training artifacts.

\subsection{Command-Provenance Trace}
\label{sec:command-trace}

The explicit proposal-to-command chain enables direct auditing of
the command actually issued by the controller. At each step, a
read-only trace stores
$ T_t = \bigl( o_t, \widetilde{o}_t,
        \phi_t, \{b_t^k\}_{k\in\Phi},
        k_t, b_t, \bar b_t, a_t, s_t, \phi_{t+1}, q_t \bigr)$,
where $k_t\in\Phi$ is the logged index of the selected phase head.
Under the hard-routing rule,
$k_t=\phi_t.$ The variable $\phi_{t+1}$ is the post-action phase, and $q_t$
contains the frozen command configuration, state-dependent bounds,
and per-rule metadata required for independent replay. Logging is observational only: it does not modify the observation,
proposal, regulator, contract, decoder, phase transition, or
returned command.

\textbf{Exact command replay.} An independent audit recomputes the full chain $b_t \longrightarrow \bar b_t \longrightarrow a_t \longrightarrow
s_t$ from the serialized observation, phase, frozen checkpoint, normalization, and command configuration.

Stored intermediate quantities are used only as comparison targets
and are not supplied as inputs to later replay stages. This tests
whether the recorded physical command is computationally identical
to the command produced by the declared policy, routing, regulation,
contract, and decoder.

\textbf{Controlled one-rule interventions.} For a qualifying pre-action state, the audit recomputes the command
while disabling exactly one deterministic mechanism. The
observation, phase, neural proposal, and all other mechanisms remain
fixed. Binding interventions test whether removing the rule produces its
predeclared local directional effect, whereas nonbinding and
inapplicable cases test that removing the rule leaves the command
unchanged.

\section{Experiments and Results}
\label{sec:experiments}
\label{sec:results}

We evaluate the controller through four complementary studies: end-to-end
closed-loop performance, matched structural ablations, command-provenance
auditing, and embodiment-specific kinematic replay. To avoid repeating
protocols and findings across separate sections, each study below presents its
setup together with its results.

\subsection{Task-Space Benchmark and Evaluation}
\label{sec:evaluation-protocol}
\label{sec:evaluation-metrics}

\textbf{Benchmark instantiation.}
Training, task-space evaluation, ablations, and trace analysis use a
project-specific deterministic analytical retinal-injection environment that
directly updates needle pose and procedural state. Targets are sampled from a
$13^{\circ}$--$20^{\circ}$ annulus about the posterior eye axis, and the
planned insertion direction has $60^{\circ}$ elevation with uniformly sampled
azimuth. Rejection sampling retains geometries whose entry lies in the
prescribed pars-plana sector, whose entry-to-target path clears the lens by at
least $1.5\,\mathrm{mm}$ and the optic-exclusion boundary by at least
$2^{\circ}$, and whose insertion axis lies within $35^{\circ}$ of the inward
surface normal. These are benchmark design parameters, not clinical safety
tolerances. Appendix Figure~\ref{fig:benchmark-sampling} summarizes the
sampling geometry and exclusion criteria. The pre-entry waypoint lies $5\,\mathrm{mm}$ outside the scleral entry. At
reset, the tip is placed a further $3$--$5\,\mathrm{mm}$ outward along
$-\boldsymbol{u}^{\star}$, giving an initial axial distance of
$8$--$10\,\mathrm{mm}$ from the entry. Two lateral offsets are sampled
independently within $\pm1\,\mathrm{mm}$ along an orthogonal basis, and initial
pitch and yaw offsets within $\pm20^{\circ}$. Episodes last at most 240 steps.
A unit Phase-A action corresponds to $500\,\mu\mathrm{m}$ of translation or
$2^{\circ}$ of angular change. Axis locking uses a
$250\,\mu\mathrm{m}$ waypoint-distance gate, a $1.5^{\circ}$ aim-error gate,
and a $25\,\mu\mathrm{m}$ entry-line tolerance. Phase A2 advances by at most
$500\,\mu\mathrm{m}$ per step, while Phase B caps physical feed at
$150\,\mu\mathrm{m}$ per step. Terminal regulation begins below
$100\,\mu\mathrm{m}$ of remaining depth, and the decoder produces no forward
motion inside the $35\,\mu\mathrm{m}$ terminal region. Robot-specific
kinematics are introduced only in the final study; robot dynamics and tissue
interaction are not modeled.

\textbf{Evaluation blocks and checkpoint lineages.}
The primary comparison uses three non-overlapping paired blocks
(E1--E3), each containing 500 held-out tasks. Each method uses one
pre-specified frozen checkpoint, and we report the mean and sample standard
deviation across blocks. These deviations reflect variation across held-out
geometry blocks rather than independently trained seeds. The matched structural study uses three additional paired 500-task blocks
(T3E1--T3E3) and five frozen checkpoints from four formal training lineages
under matched task, data, epoch, and teacher-query budgets. Because one
pre-specified training seed is used, comparisons are restricted to this
matched family. The no-correction checkpoint is the exact nominal anchor of
the Full lineage before learner-state correction, and Full continues repair from that anchor. The confirmatory trace audit uses a separate 500-episode block. The kinematic study independently replays all Flow-qualified E1--E3 trajectories, yielding
469, 475, and 470 source trajectories, respectively. All checkpoints, thresholds, and test blocks are fixed before the corresponding held-out evaluation. \textbf{Task outcomes.} Reach A2 and Reach B indicate entry into the corresponding procedural phase.
Flow requires entry into Phase B, a locked line passing within
$25\,\mu\mathrm{m}$ of the prescribed entry, and no fatal or orientation
violation. $S@\delta$ additionally requires the final 3-D
tip-to-target error to be below
$\delta\in\{100,50,35\}\,\mu\mathrm{m}$. All rates use every episode in the
evaluation block as the denominator. The hard-violation rate (HV) is the
fraction of episodes triggering any enabled fatal-event flag.

\subsection{End-to-End Task Performance}
\label{sec:compared-methods}
\label{sec:overall-results}

\textbf{Compared methods.}
All learned methods receive the same 48-dimensional observation and expose the
same five-dimensional normalized task-space interface.
\textbf{Monolithic BC} uses a two-layer 128--128 ReLU MLP with one output head
shared across phases.
\textbf{BC-GRU} uses a single-layer 128-unit GRU trained on 10-step sequences,
with hidden state retained across phase transitions.
\textbf{ACT} predicts 10-step action chunks using a width-128, two-layer,
four-head transformer with temporal ensembling; cached chunks are cleared at
phase transitions.
\textbf{Monolithic Correction IL} augments Monolithic BC with analytic-expert
labels at learner-reached states, without teacher access at evaluation.
\textbf{Correction IL + contract/decoder} applies the phase-conditioned action
contract and physical decoder to the same frozen Correction IL network, without
the learned terminal regulator. This same-checkpoint pair isolates
deployment-time action semantics and command formation.
\textbf{Ours} combines a shared encoder, hard-routed phase heads, terminal
regulation, the executable action contract, phase-dependent decoding, and
failure-focused phase-local repair.

\begin{table*}[t]
\centering
\caption{Full-task performance across three paired held-out blocks of 500
geometries each. Values are mean $\pm$ sample standard deviation across blocks;
each method uses one frozen checkpoint, so deviations reflect evaluation
rather than training variation.}
\label{tab:overall}
\footnotesize
\setlength{\tabcolsep}{2.5pt}
\resizebox{\textwidth}{!}{%
\begin{tabular}{@{}l c c c c c c c@{}}
\toprule
Method & Reach A2 (\%) $\uparrow$ & Reach B (\%) $\uparrow$ &
Flow (\%) $\uparrow$ & $S@100$ (\%) $\uparrow$ &
$S@50$ (\%) $\uparrow$ & $S@35$ (\%) $\uparrow$ &
HV (\%) $\downarrow$\\
\midrule
Monolithic BC
& $52.20\pm0.69$ & $17.47\pm2.01$ & $1.20\pm0.35$
& $1.20\pm0.35$ & $1.20\pm0.35$ & $1.20\pm0.35$
& $16.27\pm2.00$\\
BC-GRU
& $0.20\pm0.00$ & $0.00\pm0.00$ & $0.00\pm0.00$
& $0.00\pm0.00$ & $0.00\pm0.00$ & $0.00\pm0.00$
& $9.13\pm0.50$\\
ACT
& $9.93\pm0.81$ & $0.07\pm0.12$ & $0.00\pm0.00$
& $0.00\pm0.00$ & $0.00\pm0.00$ & $0.00\pm0.00$
& $40.00\pm1.06$\\
Monolithic Correction IL
& $90.47\pm0.81$ & $15.53\pm1.27$ & $12.27\pm1.45$
& $11.87\pm1.27$ & $11.73\pm1.15$ & $11.73\pm1.15$
& $3.27\pm1.45$\\
\shortstack[l]{Correction IL +decoder}
& $90.47\pm0.81$ & $90.47\pm0.81$ & $90.47\pm0.81$
& $87.93\pm1.75$ & $\mathbf{85.47\pm2.72}$
& $\mathbf{64.80\pm1.78}$ & $\mathbf{0.00\pm0.00}$\\
\textbf{Ours (structured)}
& $\mathbf{94.27\pm0.64}$ & $\mathbf{94.27\pm0.64}$
& $\mathbf{94.27\pm0.64}$ & $\mathbf{90.93\pm0.64}$
& $80.27\pm0.64$ & $64.40\pm2.43$
& $\mathbf{0.00\pm0.00}$\\
\bottomrule
\end{tabular}%
}
\end{table*}

\begin{figure}[t]
\centering
\includegraphics[height=0.15\textheight]{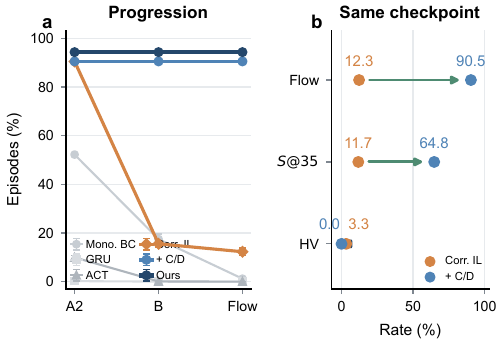}
\caption{Full-task performance. \textbf{(a)} Progression through A2, B, and
Flow. \textbf{(b)} Same-checkpoint effect of adding the contract/decoder to
Correction IL. Error bars show s.d. across three blocks; C/D denotes
contract/decoder and HV denotes hard-violation rate.}
\label{fig:main-evidence}
\end{figure}

Ours reaches $94.27\pm0.64\%$ Flow and $64.40\pm2.43\%$ $S@35$ with no hard
violations. Monolithic Correction IL reaches only $12.27\pm1.45\%$ Flow and
$11.73\pm1.15\%$ $S@35$. Applying the contract and decoder to the same frozen
Correction IL network increases Flow by 78.20 percentage points and $S@35$ by
53.07 points, while reducing HV from $3.27\%$ to zero. This same-checkpoint
comparison shows that learner-state coverage alone does not explain the
performance gap; explicit executable action semantics and phase-dependent
decoding account for most of the improvement.

Relative to Correction IL + contract/decoder, Ours improves Flow by 3.80 points
and $S@100$ by 3.00 points, but is lower by 5.20 points at $S@50$ and
0.40 points at $S@35$. Thus, the additional routed, regulated, and locally
repaired structure improves procedural coverage and the looser endpoint tier,
while the two structured deployments remain comparable at the strictest tier.
The results do not imply that every added component improves every precision
threshold. BC-GRU and ACT achieve zero Flow, while Monolithic BC reaches only
$1.20\pm0.35\%$.
\subsection{Matched Structural Ablations}
\label{sec:matched-ablations}
\label{sec:structural-results}
\begin{figure}[t]
\centering
\includegraphics[width=\columnwidth]{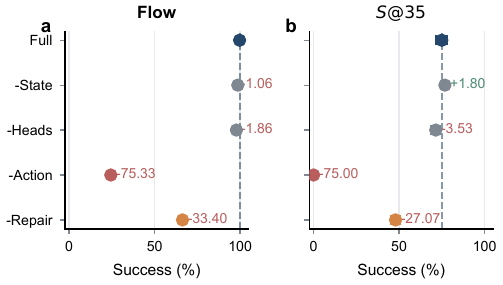}
\caption{Matched full-minus-one ablations for \textbf{(a)} Flow and
\textbf{(b)} $S@35$. Error bars show s.d. across three blocks; annotations
show percentage-point changes from Full.}
\label{fig:ablation}
\end{figure}
The matched full-minus-one family contains four variants:
\textbf{w/o derived state} masks relational task, safety, and progress fields
after standardization while retaining phase indicators and raw tip, axis,
entry, target, planned-axis, locked-axis, and waypoint geometry;
\textbf{w/o phase-conditioned heads} replaces the three independently learned
phase heads with one five-dimensional head shared across all phases;
\textbf{w/o phase-conditioned action structure} replaces scalar axial-feed
semantics, locked-axis channel binding, the A2 feed floor, the Phase-B feed
cap, terminal feed regulation, and phase-dependent decoding with unified 3T2R
expert targets, action semantics, and decoding; and
\textbf{w/o closed-loop correction} evaluates the exact nominal anchor of the
Full lineage before learner-state rollout and phase-local repair. The third
variant is deliberately an \emph{action-structure} ablation because it removes
the combined phase-dependent target semantics, deterministic contract,
regulation, and decoder rather than only one clipping operation. All variants
are frozen before the common held-out blocks; the comparison is therefore
within this matched family and does not quantify variation across random
initializations. The dominant removal is the phase-conditioned action structure. It reduces
Flow from $99.73\pm0.12\%$ to $24.40\pm0.72\%$ and drives all three endpoint
success rates to zero. The no-correction anchor shows the next-largest change,
with Flow falling by 33.40 points and $S@35$ by 27.07 points relative to Full.
These two comparisons support complementary roles: the executable action
structure determines whether phase-specific proposals can become valid physical
commands, while closed-loop correction improves proposal quality on
learner-reached failures that remain inside that structure.

Replacing the three phase heads with one shared head produces a smaller decrease
of 1.87 Flow points and 3.53 $S@35$ points. Masking derived state decreases Flow
by 1.07 points but increases $S@35$ by 1.80 points on these blocks. Given the
single training seed and the mixed direction of this change, the present study
does not establish a strict endpoint benefit from the derived-state fields.
Likewise, the routed-head result supports a modest stage-isolation benefit, not
the claim that hard routing is the primary source of the overall gain. The
largest and most stable descriptive effects in this family arise from the
phase-conditioned action structure and closed-loop correction.
Appendix Table~\ref{tab:ablation-detailed} reports the complete stage-reach,
endpoint-threshold, and hard-violation results for every matched configuration.

\subsection{Command-Provenance Trace Audit}
\label{sec:trace-results}

\begin{figure}[t]
\centering
\includegraphics[width=\columnwidth]{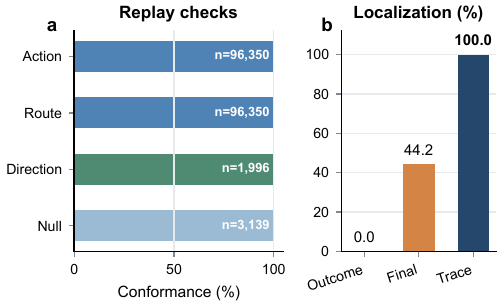}
\caption{Command-trace audit. \textbf{(a)} Replay and conformance checks.
\textbf{(b)} Violation localization using richer evidence.}
\label{fig:trace}
\end{figure}

\textbf{Protocol.}
The confirmatory log contains 96,350 official policy steps and 473 separately
recorded terminal probes. Exact replay recomputes the routed proposal,
regulated proposal, contract-compliant action, decoded command, and phase
transition from the serialized observation, phase, frozen checkpoint,
normalization, and command configuration; stored intermediate actions are used
only as comparison targets. For each mechanism and eligibility category, the
audit clones the first qualifying pre-action state in an episode and executes a
factual rule-on command and a paired one-rule-off command. The five audited
mechanisms are terminal residual regulation, locked-axis zeroing, the Phase-B
feed cap, the terminal stop band, and the A2 feed floor. Binding cases test a
predeclared directional effect, while nonbinding and inapplicable cases are
null controls. A separate synthetic test injects one trace-contract violation
and compares localization from outcome-only, final-action, and full-trace evidence. Outcome-level success cannot determine whether a stored trace follows the
controller's actual forward computation. Figure~\ref{fig:trace} therefore
audits the frozen proposal-to-command chain on this separate confirmatory
block.

Every official command is reconstructed exactly, and every routed head,
decoder, and phase transition conforms to the declared execution order. All
1,996 binding one-rule-off pairs move in their predeclared direction, while all
3,139 null controls remain unchanged. The measured local effects are also
consistent with the intended mechanisms: disabling terminal residual
regulation raises normalized feed by 0.00328; disabling the Phase-B feed cap
raises it by 0.27944; disabling the A2 floor lowers it by 0.69735; restoring
channels removed by locked-axis zeroing produces a non-feed $L_{\infty}$ change
of 0.00993 while decoded axial motion remains zero; and disabling the stop band
adds $0.341\,\mu\mathrm{m}$ of axial motion.

For the 2,496 injected single-rule violations, outcome-only evidence provides
no localization, final-action evidence reaches 44.2\%, and the full trace
reaches 100.0\%. This result is confined to the evaluated synthetic violation
family. Together, the audit establishes computational faithfulness of the
recorded task-space command chain and the local directional role of its
deterministic mechanisms. It does not interpret hidden neural representations,
estimate the episode-level causal effect of permanently disabling a rule, or
validate physical or clinical safety.
Appendix Table~\ref{tab:trace-detailed} separates the reconstruction,
transition, directional-intervention, null-control, and synthetic-localization
checks together with their distinct evaluation-unit counts.

\subsection{Kinematic Compatibility}
\label{sec:kinematic-results}

\textbf{Protocol.}
Current RCM is our project-specific analytical model. The Franka Panda,
KUKA LBR iiwa 14 R820, and UR5 kinematic descriptions and joint limits are
taken from pinned snapshots of the corresponding Franka Robotics, lbr-stack,
and Universal Robots model repositories
\citep{FrankaROS2025,LBRDescription2026,URDescription2026}. Kinova Mico is extracted from the bundled
\texttt{Mico arm} model in CoppeliaSim 4.10.0. Each of
these five embodiments receives the same frozen task-space trajectories
independently. IK begins from
the first task-space pose; the robot home configuration is used only as the
initial IK guess. Targets are processed in temporal order, and replay stops at
the first IK, FK, joint-limit, or continuity failure. Each embodiment uses one
fixed robot-to-task transform. We report start reachability, full replay,
overall success, accepted-step coverage, position and axis errors, and
joint-limit/continuity failures. CoppeliaSim 4.10.0 rev0
\citep{Rohmer2013} is used for frame and tool-transform checks and synchronized
visualization of selected traces, not for policy training or dynamic control
evaluation. The policy, observation
semantics, phase logic, and decoded task-space commands remain fixed; only the
robot model, frames, joint limits, and IK/FK adapter change.

\begin{figure}[t]
\centering
\includegraphics[width=\columnwidth]{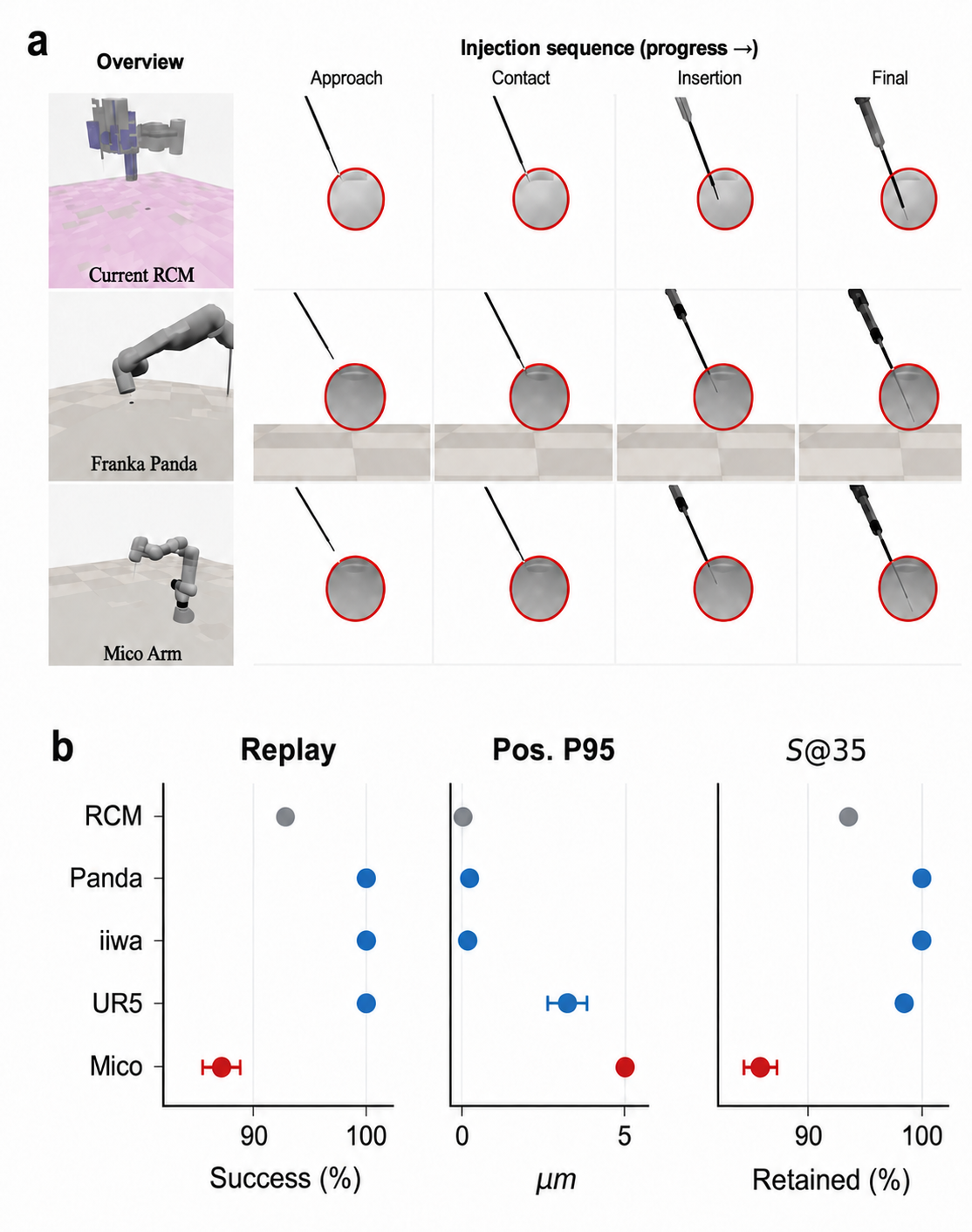}
\caption{Kinematic compatibility across robot embodiments.
\textbf{(a)} Representative CoppeliaSim sequences.
\textbf{(b)} Replay success, position P95 error, and retained $S@35$.
Error bars show s.d. across E1--E3.}
\label{fig:kinematics}
\end{figure}

Franka Panda and UR5 achieve $100.00\pm0.00\%$ overall replay success, and
KUKA LBR iiwa reaches $99.86\pm0.12\%$. Current RCM reaches
$92.79\pm0.74\%$ overall; its lower value is primarily attributable to start
reachability ($93.07\pm0.66\%$), because full replay succeeds on
$99.70\pm0.13\%$ of reachable starts. Kinova Mico has the lowest overall
success ($87.48\pm1.60\%$) and the highest joint-limit/continuity failure rate
($4.14\pm1.02\%$). Across all embodiments, position P95 is at most
$5.11\,\mu\mathrm{m}$ and axis P95 is at most
$1.869\times10^{-3}\,{}^{\circ}$.

Endpoint $S@35$ preservation ranges from $85.60\pm1.06\%$ for Kinova Mico to
$100.00\pm0.00\%$ for Franka Panda. These results show that the frozen
task-space interface is numerically realizable through several distinct
kinematic adapters, but they do not evaluate command-conditioned robot
tracking, dynamics, contact, tissue interaction, sensing uncertainty, or
multi-robot cooperation.
Appendix Table~\ref{tab:kinematics-detailed} provides the complete
embodiment-level replay, conditioning, precision, and endpoint-preservation
values.

\section{Conclusion}
We present a safety-structured task-space policy for staged microsurgical
injection. Across three held-out blocks, it achieves
$94.27\pm0.64\%$ Flow and $64.40\pm2.43\%$ $S@35$ with no hard violations;
ablations identify phase-dependent action semantics as critical, while the
trace exactly reconstructs 96,350 commands and validates 1,996 eligible rule
interventions. These results support auditable staged control through explicit
action semantics.

\clearpage
\bibliography{sample-base}
\clearpage

\FloatBarrier
\appendix

\section{Benchmark Environment and Evaluation Protocol}
\label{app:benchmark-environment}

This section provides the complete task definition underlying the
task-space experiments in the main paper. It distinguishes the analytical
benchmark used for policy training and closed-loop evaluation from the
embodiment-specific kinematic replay reported separately below. Unless
otherwise stated, policy training, learner-state reconstruction, task-space
evaluation, structural ablation, and command-trace analysis are conducted in
the same frozen analytical environment.

\subsection{Analytical Task-Space Environment}

The benchmark is a project-specific deterministic analytical simulator of a
staged retinal-injection procedure. It directly represents the needle tip,
needle axis, prescribed scleral entry, subretinal target, planned insertion
axis, accepted locked axis, and current procedural phase. At each control step,
the policy receives the current observation and produces a normalized
five-dimensional output. The phase-conditioned command-generation chain
converts this output into a physical task-space command, after which the
analytical environment updates the needle pose and procedural state.

The benchmark operates directly in needle task space rather than through a
particular robot embodiment. Policy inference is therefore not affected by
robot-specific inverse kinematics, joint limits, tracking error, or robot
dynamics. Robot models are introduced only in the independent kinematic replay
reported in Section~\ref{app:kinematic-details}. This separation prevents
embodiment-specific IK failure from being interpreted as failure of the learned
task-space policy.

The simulator does not model tissue deformation, needle--tissue interaction,
force feedback, actuator dynamics, command-tracking error, sensing uncertainty,
or clinical complications. It evaluates staged task-space command semantics,
closed-loop geometric progression, endpoint precision, and declared logical
violations. The results should therefore not be interpreted as complete
physical or clinical validation.

\subsection{Observation and Normalized Action Interface}

All learned methods receive the same $48$-dimensional observation and expose
the same five-dimensional normalized task-space interface. The observation
contains the task geometry and procedural state required by the controller,
including the current needle tip and axis, prescribed entry and target,
planned and locked axes, waypoint geometry, phase information, and derived
relational, safety, and progress variables.

The five-dimensional neural output is deliberately fixed across phases so that
the evaluation isolates how different methods handle changing action
semantics. For monolithic baselines, one shared output mapping is used
throughout the procedure. For the structured policy, the same normalized
coordinate dimension is retained, but phase routing, deterministic
admissibility constraints, and phase-dependent decoding determine which
coordinates are executable and what physical meanings they receive.

The five normalized coordinates should therefore not be interpreted as five
fixed physical degrees of freedom throughout the entire episode. During
pre-aim, the coordinates support translational and orientational correction.
After the insertion line is accepted, lateral and angular coordinates are
disabled and the remaining active coordinate is interpreted as progress along
the locked insertion line.

\subsection{Procedural Phases}

The benchmark contains three procedural phases:
\begin{equation}
    \Phi
    =
    \{
        \mathrm{A},
        \mathrm{A2},
        \mathrm{B}
    \}.
\end{equation}

\paragraph{Phase $\mathrm{A}$: pre-aim.}
The controller moves the needle toward a pre-entry waypoint and aligns the
current needle line with the planned insertion direction. Translational and
orientational correction channels remain available. A unit normalized
Phase-$\mathrm{A}$ action corresponds to $500\,\mu\mathrm{m}$ of translation
or $2^{\circ}$ of angular change in the associated coordinate.

\paragraph{Phase $\mathrm{A2}$: guarded entry.}
Once the waypoint and alignment conditions are satisfied, the current needle
line is accepted as a geometric commitment. Lateral and angular motion are
removed from the executable action set, and the remaining active coordinate
advances the needle along the locked line. The Phase-$\mathrm{A2}$ contract
introduces a positive progress floor and permits at most
$500\,\mu\mathrm{m}$ of physical progress per step.

\paragraph{Phase $\mathrm{B}$: insertion.}
After entry, the controller continues along the same locked line toward the
target. Physical feed is capped at $150\,\mu\mathrm{m}$ per step. Terminal
feed regulation begins when the remaining insertion depth falls below
$100\,\mu\mathrm{m}$, and the decoder produces no additional forward motion
inside the final $35\,\mu\mathrm{m}$ stop region.

These phase definitions form part of the frozen benchmark configuration. They
are not inferred from demonstrations and are not changed between methods or
evaluation blocks.

\subsection{Initialization and Phase Transitions}

The pre-entry waypoint lies $5\,\mathrm{mm}$ outside the prescribed scleral
entry along the negative planned insertion direction. At reset, the needle tip
is placed a further $3$--$5\,\mathrm{mm}$ outward along the same direction,
giving an initial axial distance of $8$--$10\,\mathrm{mm}$ from the entry.

Two independent lateral offsets are sampled within
$\pm1\,\mathrm{mm}$ along an orthogonal basis around the planned axis.
Initial pitch and yaw perturbations are independently sampled within
$\pm20^{\circ}$. These perturbations require the policy to correct both
position and orientation before line acceptance.

Transition from Phase $\mathrm{A}$ to Phase $\mathrm{A2}$ uses the following
frozen gates:
\begin{itemize}
    \item waypoint distance no greater than $250\,\mu\mathrm{m}$;
    \item aim error no greater than $1.5^{\circ}$; and
    \item entry-line error no greater than $25\,\mu\mathrm{m}$.
\end{itemize}

Each episode lasts at most $240$ control steps. Phase transitions are produced
by the benchmark task logic after applying the physical task-space command;
they are not selected directly by the neural policy.

\subsection{Outcome Definitions}

We report both stage-progression and endpoint metrics.

\paragraph{Reach A2 and Reach B.}
\emph{Reach A2} indicates that the episode satisfies the pre-aim acceptance
conditions and enters guarded entry. \emph{Reach B} indicates that the episode
subsequently enters the insertion phase.

\paragraph{Flow.}
An episode satisfies \emph{Flow} only if it enters Phase $\mathrm{B}$, retains
a locked line passing within $25\,\mu\mathrm{m}$ of the prescribed entry, and
produces no fatal or orientation violation.

\paragraph{Endpoint success.}
For $\delta\in\{100,50,35\}\,\mu\mathrm{m}$, $S@\delta$ additionally requires
the final three-dimensional tip-to-target error to be below $\delta$. Every
stage and endpoint rate uses all episodes in the evaluation block as its
denominator; endpoint success is not conditioned on Flow.

\paragraph{Hard violations.}
The hard-violation rate, denoted HV, is the fraction of episodes in which any
enabled fatal-event flag is triggered. It is also computed over all episodes
in the corresponding evaluation block.

\subsection{Evaluation Blocks and Frozen Checkpoints}

The primary end-to-end comparison uses three non-overlapping paired blocks,
each containing $500$ held-out geometries:
\begin{align*}
    \mathrm{E1} &: \text{seeds }100\text{--}599,\\
    \mathrm{E2} &: \text{seeds }600\text{--}1099,\\
    \mathrm{E3} &: \text{seeds }1100\text{--}1599.
\end{align*}

Every method uses one pre-specified frozen checkpoint. Reported standard
deviations therefore quantify variation across held-out geometry blocks rather
than variation across independently trained random seeds.

The matched structural study uses three additional paired $500$-geometry
blocks:
\begin{align*}
    \mathrm{T3E1} &: \text{seeds }177000\text{--}177499,\\
    \mathrm{T3E2} &: \text{seeds }177500\text{--}177999,\\
    \mathrm{T3E3} &: \text{seeds }178000\text{--}178499.
\end{align*}

These blocks evaluate five frozen checkpoints from four formal training
lineages under matched task, data, epoch, and teacher-query budgets. One
pre-specified training seed is used, so the comparison is confined to this
matched structural family.

The confirmatory command-trace audit uses a separate $500$-episode block with
seeds $931000$--$931499$. The embodiment-specific study independently replays
the Flow-qualified trajectories from E1--E3, yielding $469$, $475$, and $470$
source trajectories, respectively. No checkpoint is retrained using any
evaluation block.

\section{Baseline Implementations and Comparison Controls}
\label{app:baseline-details}

All learned methods receive the same $48$-dimensional observation, expose the
same five-dimensional normalized policy interface, and are evaluated on the
same paired held-out geometries. No method receives online expert access during
evaluation.

The baselines differ in temporal representation, learner-state coverage, and
treatment of phase-dependent action semantics. The comparison addresses three
questions:
\begin{enumerate}
    \item whether a monolithic feed-forward policy can learn the complete
    staged behavior from offline supervision;
    \item whether recurrence, action chunking, or learner-state correction
    resolves the main closed-loop failures; and
    \item whether an explicit phase-conditioned contract and physical decoder
    remain important when the underlying neural checkpoint is held fixed.
\end{enumerate}

\subsection{Monolithic Behavioral Cloning}

Monolithic BC uses a two-layer multilayer perceptron with hidden widths
$128$--$128$ and ReLU activations. A single five-dimensional output head is
shared across all phases. Phase and task state are available through the
observation, but there is no phase-specific executable routing, deterministic
channel binding, or phase-dependent physical decoder.

This baseline tests whether a standard feed-forward imitation-learning policy
can infer both the desired command and its changing phase semantics entirely
from supervised data.

\subsection{BC-GRU}

BC-GRU replaces the feed-forward MLP with a single-layer GRU containing
$128$ hidden units. It is trained on sequences of length $10$, and its hidden
state is retained across procedural phase transitions.

This baseline tests whether short temporal context and recurrent memory are
sufficient to resolve the staged task. Although the hidden state can retain
procedural history, BC-GRU continues to use a shared output interpretation and
does not apply an explicit phase-conditioned action contract.

\subsection{Action Chunking Transformer}

ACT predicts chunks of $10$ future actions using a transformer with width
$128$, two transformer layers, and four attention heads. Temporal ensembling
combines overlapping chunk predictions. Cached chunks are cleared at
procedural phase transitions so that commands generated under a previous phase
are not executed after the transition.

ACT tests whether short-horizon prediction and temporal ensembling improve
closed-loop stability. Clearing cached chunks prevents stale commands from
crossing phase boundaries, but newly predicted outputs still use a monolithic
action interface without the proposed deterministic contract and decoder.

\subsection{Monolithic Correction IL}

Monolithic Correction IL begins from the Monolithic BC checkpoint and augments
training with analytic-expert targets at states reached by the learner. Here,
IL denotes imitation learning. The additional data improve coverage of the
learner-induced state distribution, but the policy continues to use one shared
five-dimensional output mapping across all phases.

The analytic expert is used only to generate correction labels during
training. It is not queried during held-out evaluation. This baseline therefore
tests whether learner-state correction alone can resolve the closed-loop
failures of behavioral cloning.

\subsection{Correction IL with Contract and Decoder}

Correction IL + contract/decoder uses exactly the same frozen neural checkpoint
as Monolithic Correction IL. No network parameter is retrained or otherwise
changed. Instead, the five-dimensional output is passed through the
phase-conditioned executable action contract and phase-dependent physical
decoder.

This condition omits the learned terminal regulator used by the complete
method. It is therefore not the full proposed controller. Its purpose is to
isolate the deployment-time effect of explicit action admissibility and
phase-dependent physical command formation while holding the neural proposal
network fixed.

Because the two Correction IL conditions use the same checkpoint, differences
between them cannot be attributed to network capacity, optimizer behavior,
learner-state labels, or training duration.

\subsection{Structured Policy}

The complete structured policy combines:
\begin{enumerate}
    \item a shared observation encoder;
    \item hard-routed phase-specific proposal heads;
    \item state-dependent terminal feed regulation;
    \item the phase-conditioned executable action contract;
    \item deterministic phase-dependent physical decoding; and
    \item failure-focused phase-local repair.
\end{enumerate}

Only the proposal-generation components are learned. Channel masks,
admissible bounds, locked-axis semantics, and physical decoding rules are
provided by the frozen task specification. During phase-local repair, only the
pre-aim proposal head is updated; the shared encoder, guarded-entry head,
insertion head, and terminal regulator remain fixed.

\subsection{Scope of the Baseline Comparison}

The monolithic baselines test whether increased temporal modeling or improved
learner-state coverage can compensate for an implicit action contract. The
same-checkpoint Correction IL comparison addresses a narrower question:
whether explicitly changing how a fixed neural output is admitted and decoded
can alter closed-loop task completion.

The comparison does not claim that these architectures exhaust all recurrent,
chunked, safety-filtered, or hierarchical controllers. It instead controls the
observation and normalized output interface while progressively introducing
temporal memory, learner-state correction, and executable phase semantics.

\section{Benchmark Geometry and Sampling Constraints}
\label{app:benchmark-geometry}

\begin{figure}[t]
\centering
\includegraphics[width=0.90\columnwidth]{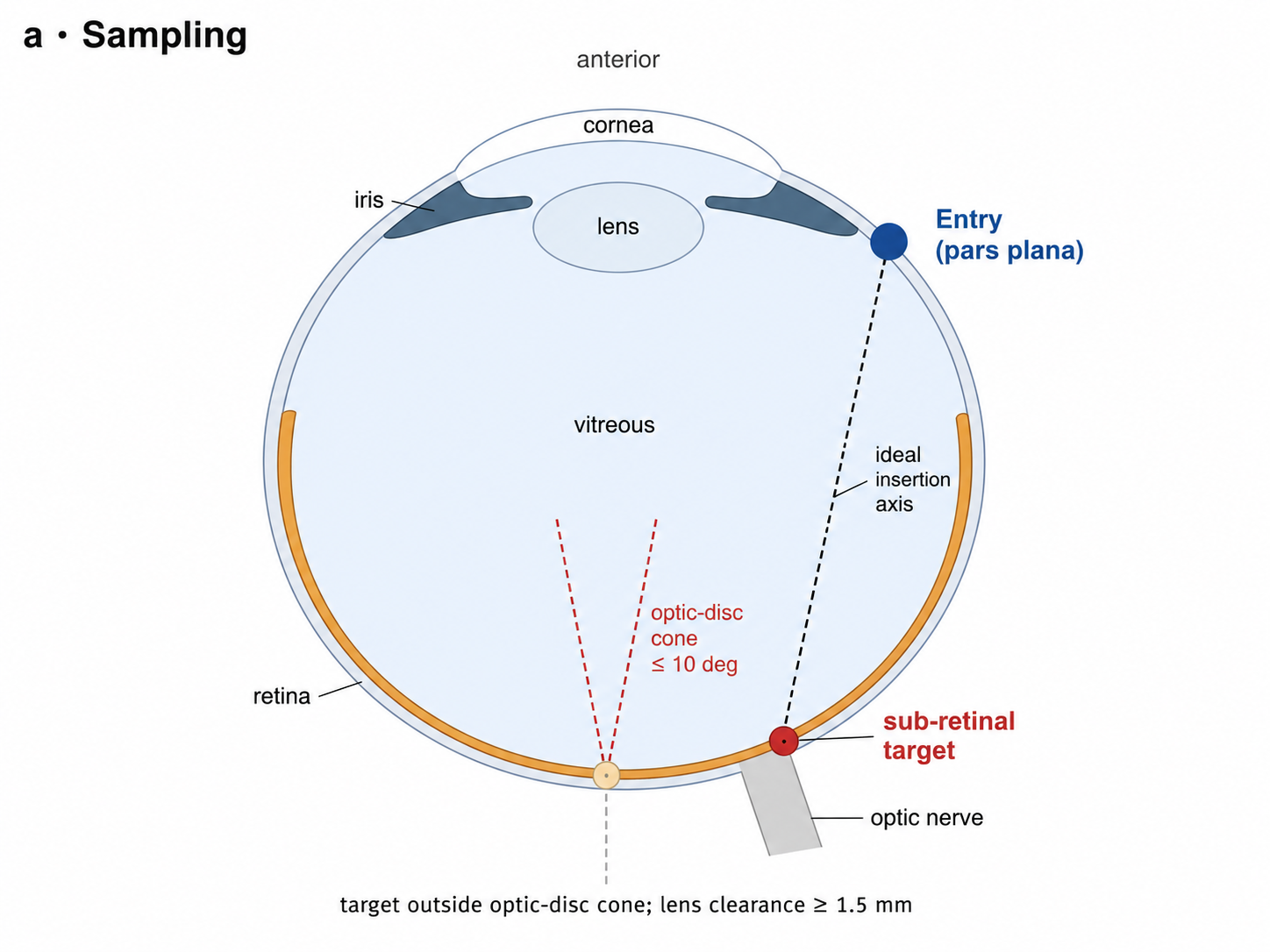}
\caption{Benchmark sampling geometry. The dashed entry--target line is the
planned insertion axis. Retained paths clear the lens by at least
$1.5\,\mathrm{mm}$ and remain at least $2^{\circ}$ outside the illustrated
$10^{\circ}$ optic-exclusion cone. The planned line must additionally lie
within $35^{\circ}$ of the inward surface normal. Schematic, not to scale.}
\label{fig:benchmark-sampling}
\end{figure}

Figure~\ref{fig:benchmark-sampling} maps the analytical task geometry. The eye,
entry, target, planned insertion line, lens exclusion region, and optic
exclusion region are represented directly in the benchmark coordinate frame.
Sampling is performed before policy execution, and candidate geometries are
retained only if all predeclared geometric conditions are satisfied.

Targets are sampled from a $13^{\circ}$--$20^{\circ}$ annulus about the
posterior eye axis. Candidate insertion directions have fixed
$60^{\circ}$ elevation and uniformly sampled azimuth. Rejection sampling then
retains only geometries that:
\begin{enumerate}
    \item place the entry in the prescribed pars-plana sector;
    \item clear the lens by at least $1.5\,\mathrm{mm}$;
    \item remain at least $2^{\circ}$ outside the boundary of the
    $10^{\circ}$ optic-exclusion cone; and
    \item keep the planned line within $35^{\circ}$ of the inward surface
    normal at the entry.
\end{enumerate}

These filters depend only on the sampled task geometry and not on the evaluated
policy. Consequently, paired methods receive the same retained geometries
within each evaluation block.

The listed values are simulation design parameters selected to define the
analytical benchmark. They are not asserted clinical safety tolerances,
surgical planning recommendations, or validated anatomical limits.

\section{Detailed Experimental Tables}
\label{app:detailed-tables}

This appendix provides the complete numerical records underlying
Figures~\ref{fig:ablation}--\ref{fig:kinematics}. It exposes metrics omitted
from the compact single-column visualizations, defines the evaluation unit and
conditioning associated with each percentage, and makes the scope of each
validation explicit.

No additional training runs, held-out geometries, trace cases, or robot
trajectories are introduced here.
Tables~\ref{tab:ablation-detailed}--\ref{tab:kinematics-detailed} reorganize
the same evaluations reported in the main text and therefore provide expanded
reporting rather than an independent replication.

\subsection{Matched Structural Ablations}

Table~\ref{tab:ablation-detailed} expands Figure~\ref{fig:ablation} from Flow
and $S@35$ to the complete stage-progression sequence, three endpoint
tolerances, and hard-violation rate. Every configuration is frozen before
evaluation on the common T3E1--T3E3 blocks.

\paragraph{Full.}
The complete configuration includes the shared encoder, phase-conditioned
proposal heads, derived observation state, terminal feed regulation,
executable action contract, phase-dependent decoder, and closed-loop
phase-local correction.

\paragraph{Without derived state.}
This variant masks relational task, safety, and progress fields after
standardization while retaining phase indicators and the raw tip, axis, entry,
target, planned-axis, locked-axis, and waypoint geometry. It tests whether the
derived relational variables provide information beyond the underlying
geometric quantities.

\paragraph{Without phase-conditioned heads.}
The independently learned phase heads are replaced by one five-dimensional
head shared across all phases. The deterministic contract and
phase-dependent decoder remain present. This variant isolates the contribution
of learned phase-specific proposal mappings.

\paragraph{Without phase-conditioned action structure.}
This variant removes scalar axial-feed semantics, locked-axis channel binding,
the Phase-$\mathrm{A2}$ feed floor, Phase-$\mathrm{B}$ feed cap, terminal
feed regulation, and phase-dependent decoding. These components are replaced
with unified 3T2R expert targets, shared action semantics, and unified decoding.

This is intentionally a structural ablation rather than an ablation of a
single clipping operation. It evaluates the combined role of phase-dependent
target semantics, admissibility, regulation, and physical decoding.

\paragraph{Without closed-loop correction.}
This checkpoint is the exact nominal anchor of the Full lineage before
learner-state rollout and phase-local repair. It retains the complete
structured command-generation path but has not been updated on selected
learner-reached failure states.

Because one pre-specified training seed is used, the standard deviations
quantify variation across the three paired evaluation blocks rather than
sensitivity to random initialization.

The Reach A2, Reach B, and Flow columns should be read from left to right as a
stage-progression diagnostic. Removing the phase-conditioned action structure
allows some episodes to approach A2 but sharply reduces entry into Phase B and
drives all endpoint success rates to zero. Removing closed-loop correction
causes an earlier loss in stage progression and a consistent decline across
all three endpoint tolerances.

Removing derived state or replacing the phase-specific heads produces smaller
and partly mixed changes. The table therefore supports the main-text
attribution to executable action semantics and learner-state correction while
preserving the single-seed limitation on the smaller component effects.

\begin{table*}[t]
\centering
\caption{Complete matched structural full-minus-one results underlying
Figure~\ref{fig:ablation}. Values are mean $\pm$ sample standard deviation
across three paired held-out blocks of 500 geometries each. One pre-specified
training seed is used.}
\label{tab:ablation-detailed}
\footnotesize
\setlength{\tabcolsep}{2.5pt}
\resizebox{\textwidth}{!}{%
\begin{tabular}{@{}l c c c c c c c@{}}
\toprule
Configuration & Reach A2 (\%) $\uparrow$ & Reach B (\%) $\uparrow$ &
Flow (\%) $\uparrow$ & $S@100$ (\%) $\uparrow$ &
$S@50$ (\%) $\uparrow$ & $S@35$ (\%) $\uparrow$ &
HV (\%) $\downarrow$\\
\midrule
\textbf{Full}
& $\mathbf{99.73\pm0.12}$ & $\mathbf{99.73\pm0.12}$
& $\mathbf{99.73\pm0.12}$ & $\mathbf{96.40\pm0.40}$
& $\mathbf{89.53\pm1.47}$ & $75.00\pm3.14$
& $0.00\pm0.00$\\
w/o derived state
& $98.67\pm0.50$ & $98.67\pm0.50$ & $98.67\pm0.50$
& $95.73\pm0.70$ & $89.27\pm1.36$
& $\mathbf{76.80\pm0.92}$ & $0.00\pm0.00$\\
w/o phase-conditioned heads
& $97.87\pm0.58$ & $97.87\pm0.58$ & $97.87\pm0.58$
& $94.20\pm1.22$ & $87.60\pm1.31$ & $71.47\pm2.66$
& $0.00\pm0.00$\\
\shortstack[l]{w/o phase-conditioned\\action structure}
& $68.53\pm0.23$ & $24.47\pm0.70$ & $24.40\pm0.72$
& $0.00\pm0.00$ & $0.00\pm0.00$ & $0.00\pm0.00$
& $0.07\pm0.12$\\
w/o closed-loop correction
& $66.33\pm0.61$ & $66.33\pm0.61$ & $66.33\pm0.61$
& $63.87\pm1.30$ & $59.93\pm2.01$ & $47.93\pm2.53$
& $0.07\pm0.12$\\
\bottomrule
\end{tabular}%
}
\end{table*}

\subsection{Command-Trace Audit}

Table~\ref{tab:trace-detailed} separates the distinct audit questions
compressed into Figure~\ref{fig:trace}. The audit is conducted on a separate
confirmatory block. Stored intermediate quantities are never supplied as
inputs to later replay stages.

For exact reconstruction, the independent audit recomputes
\begin{equation}
    b_t
    \longrightarrow
    \bar b_t
    \longrightarrow
    a_t
    \longrightarrow
    s_t
\end{equation}
from the serialized observation, procedural phase, frozen checkpoint,
normalization, and command configuration. Stored proposals, actions, and
commands are used only as comparison targets.

Routing and transition checks separately verify that the selected proposal
head agrees with the current phase, command formation follows the declared
execution order, and the recorded post-action phase conforms to the frozen
transition logic.

The one-rule-off audit covers:
\begin{enumerate}
    \item terminal residual regulation;
    \item locked-axis zeroing;
    \item the Phase-$\mathrm{B}$ feed cap;
    \item the terminal stop band; and
    \item the Phase-$\mathrm{A2}$ feed floor.
\end{enumerate}

For each mechanism, the audit clones a qualifying pre-action state and computes
a factual rule-on command together with a paired command in which exactly that
mechanism is disabled. The observation, phase, neural proposal, and all other
mechanisms remain fixed.

The $1{,}996$ directional pairs include only eligible cases in which the
removed rule is binding. The $3{,}139$ null cases contain inactive,
nonbinding, or inapplicable mechanisms. These groups use different eligibility
conditions and should not be combined into one denominator.

The $2{,}496$ localization cases form a separate synthetic violation family.
Each case contains one injected trace-contract violation, and localization is
attempted using outcome-only, final-action, or full-trace evidence.

Exact reconstruction and directional agreement establish computational
conformance for the evaluated frozen command path. They do not establish
robustness to arbitrary software faults, causal interpretability of hidden
neural representations, physical robot tracking, or clinical safety.

\begin{table*}[t]
\centering
\caption{Complete command-trace reconstruction and rule-level audit results
underlying Figure~\ref{fig:trace}. The 96,350 official steps exclude 473
terminal probes logged separately.}
\label{tab:trace-detailed}
\begin{tabularx}{\textwidth}{@{}X r X@{}}
\toprule
Check & Evaluation unit ($n$) & Result\\
\midrule
Staged-action reconstruction
& 96,350 steps
& 100\%; maximum absolute action error $=0$\\
Routing and transition conformance
& 96,350 steps
& 100\%; illegal transitions $=0$\\
Expected-direction agreement across five rules
& 1,996 pairs
& 100.0\%\\
Inactive, nonbinding, or inapplicable rules
& 3,139 cases
& 100\%; no spurious effect\\
Synthetic trace-contract localization
& 2,496 cases
& Outcome/final-action/full-trace accuracy: 0.0 / 44.2 / 100.0\%\\
\bottomrule
\end{tabularx}
\end{table*}

\subsection{Embodiment-Specific Kinematic Replay}
\label{app:kinematic-details}

Table~\ref{tab:kinematics-detailed} expands Figure~\ref{fig:kinematics} for the
same Flow-qualified task-space trajectories. The policy checkpoint,
observation semantics, phase logic, action contract, and decoded commands
remain fixed. There is no policy retraining, fine-tuning, embodiment-specific
adaptation, or resampling of the source trajectories.

Five embodiments are evaluated independently: the project-specific Current
RCM model, Franka Panda, KUKA LBR iiwa 14 R820, UR5, and Kinova Mico. Each
embodiment uses one fixed robot-to-task transform and receives the same ordered
task-space poses. CoppeliaSim 4.10.0 is used for frame checks,
tool-transform verification, and synchronized visualization, not for policy
training or dynamic control evaluation.

Inverse kinematics begins from the first requested task-space pose. The robot
home configuration is used only as the initial IK guess. Subsequent target
poses are processed in temporal order, and replay stops at the first IK,
forward-kinematics, joint-limit, or continuity failure.

The metrics use the following conditioning.

\paragraph{Start.}
Start is the fraction of all source trajectories for which the first
task-space pose is reachable.

\paragraph{Replay.}
Replay is the fraction of reachable-start trajectories reproduced successfully
to completion.

\paragraph{Overall.}
Overall is the fraction of all source trajectories that have a reachable start
and complete the full replay.

\paragraph{Coverage.}
Coverage is the fraction of requested trajectory steps accepted by sequential
replay, conditioned on a reachable start.

\paragraph{Position and axis errors.}
Position P95 and Axis P95 are the $95$th percentiles of the
forward-kinematics reconstruction errors over accepted poses.

\paragraph{Joint-limit or continuity failure.}
J/C is the fraction of reachable-start trajectories terminated by a joint-limit
or continuity failure.

Failed starts and incomplete replays count as unsuccessful when computing
$S@100$, $S@50$, and $S@35$. Strict endpoint preservation is defined as
\begin{equation}
\begin{gathered}
    \text{$S@35$ preservation}\\
    =
    \frac{
        \text{source $S@35$ successes retained after replay}
    }{
        \text{source $S@35$ successes}
    }.
\end{gathered}
\end{equation}

The replay study evaluates numerical compatibility of the frozen task-space
interface with distinct kinematic adapters. It does not evaluate tracking
dynamics, actuator error, contact, tissue interaction, or sensing uncertainty.

\begin{table*}[t]
\centering
\caption{Complete embodiment-specific sequential IK/FK replay results
underlying Figure~\ref{fig:kinematics}. Values are mean $\pm$ sample standard
deviation across E1--E3. Every embodiment receives the same Flow-qualified
source trajectories independently.}
\label{tab:kinematics-detailed}
\vspace{2pt}
\textbf{(a) Kinematic replay.}

\vspace{2pt}
\footnotesize
\setlength{\tabcolsep}{1.5pt}
\resizebox{\textwidth}{!}{%
\begin{tabular}{@{}l c c c c c c c@{}}
\toprule
Robot &
Start (\%) $\uparrow$ &
Replay (\%) $\uparrow$ &
Overall (\%) $\uparrow$ &
Coverage (\%) $\uparrow$ &
Pos. P95 ($\mu\mathrm{m}$) $\downarrow$ &
Axis P95 ($\times10^{-3}\,{}^{\circ}$) $\downarrow$ &
J/C fail. (\%) $\downarrow$\\
\midrule
Current RCM
& $93.07\pm0.66$ & $99.70\pm0.13$ & $92.79\pm0.74$
& $99.70\pm0.13$ & $0.00\pm0.00$ & $0.000\pm0.000$
& $0.30\pm0.13$\\
Franka Panda
& $100.00\pm0.00$ & $100.00\pm0.00$ & $100.00\pm0.00$
& $100.00\pm0.00$ & $0.31\pm0.00$ & $0.101\pm0.002$
& $0.00\pm0.00$\\
KUKA LBR iiwa
& $99.86\pm0.12$ & $100.00\pm0.00$ & $99.86\pm0.12$
& $100.00\pm0.00$ & $0.28\pm0.02$ & $0.069\pm0.024$
& $0.00\pm0.00$\\
UR5
& $100.00\pm0.00$ & $100.00\pm0.00$ & $100.00\pm0.00$
& $100.00\pm0.00$ & $3.15\pm0.65$ & $0.378\pm0.060$
& $0.00\pm0.00$\\
Kinova Mico
& $99.01\pm0.64$ & $88.36\pm2.19$ & $87.48\pm1.60$
& $93.36\pm1.38$ & $5.11\pm0.08$ & $1.869\pm0.042$
& $4.14\pm1.02$\\
\bottomrule
\end{tabular}%
}

\vspace{3pt}
\footnotesize
Start, Replay, Overall, and Coverage denote start reachability, full replay,
overall success, and accepted-step coverage. Start and Overall use all source
trajectories; Replay, Coverage, and J/C are conditioned on a reachable start.
J/C denotes joint-limit or continuity failure.

\vspace{7pt}
\textbf{(b) Endpoint precision after replay.}

\vspace{2pt}
\resizebox{0.82\textwidth}{!}{%
\begin{tabular}{@{}l c c c c@{}}
\toprule
Robot & $S@100$ (\%) $\uparrow$ & $S@50$ (\%) $\uparrow$ &
$S@35$ (\%) $\uparrow$ & $S@35$ preservation (\%) $\uparrow$\\
\midrule
Current RCM
& $89.40\pm1.81$ & $78.93\pm1.46$ & $63.80\pm2.56$
& $93.38\pm0.24$\\
Franka Panda
& $96.47\pm1.34$ & $85.15\pm1.18$ & $68.33\pm2.88$
& $100.00\pm0.00$\\
KUKA LBR iiwa
& $96.33\pm1.21$ & $85.01\pm1.08$ & $68.18\pm2.82$
& $99.80\pm0.18$\\
UR5
& $96.47\pm1.34$ & $85.01\pm0.86$ & $67.20\pm2.96$
& $98.23\pm0.39$\\
Kinova Mico
& $84.09\pm0.42$ & $74.33\pm1.18$ & $58.56\pm2.73$
& $85.60\pm1.06$\\
\bottomrule
\end{tabular}%
}

\vspace{2pt}
\footnotesize
Failed starts or replays count as unsuccessful. $S@35$ preservation is the
fraction of source $S@35$ successes retained after replay.
\end{table*}

\FloatBarrier
\end{document}